\documentclass[11pt,a4paper]{article}
\usepackage[dvipsnames]{xcolor}
\usepackage[hyperref]{acl2020}
\usepackage{times}
\usepackage{latexsym}

\usepackage{microtype}

\aclfinalcopy 

\usepackage{graphicx}
\usepackage{amsmath}
\usepackage{amssymb}
\usepackage[shortlabels]{enumitem}
\usepackage[font={small}]{caption}
\usepackage[multiple]{footmisc}
\usepackage{multirow}
\usepackage{booktabs}
\usepackage[T1]{fontenc}
\usepackage[dvipsnames]{xcolor}

\newcommand\mg[1]{\textcolor{purple}{MG: #1}}

\newcommand\nl[1]{{\it``#1''}}
\newcommand\ssc[1]{\textsubscript{\textsc{#1}}}
\newcommand\bert{\textsc{BERT}}
\newcommand\roberta{\textsc{RoBERTa}}
\newcommand\genbert{\textsc{GenBERT}}
\newcommand\genroberta{\textsc{GenRoBERTa}}
\newcommand\addsub{\textsc{AddSub}}
\newcommand\singleop{\textsc{SOp}}

\newcommand\singleeq{\textsc{SEq}}
\newcommand\nabert{\textsc{NABERT+}}
\newcommand\mtmsn{\textsc{MTMSN}}
\newcommand\comment[1]{}
\newcommand\drop{\textsc{DROP}}

\newcommand\squad{\textsc{SQuAD}}
\newcommand\question{\mathbf{q}}
\newcommand\context{\mathbf{c   }}
\newcommand\answer{\mathbf{a}}

\definecolor{myblue}{RGB}{60,120,216}
\definecolor{myred}{RGB}{221,126,107}

\author{Mor Geva\thanks{These authors contributed equally.} \\
  Tel Aviv University, \\
  Allen Institute for AI \\
  {\tt morgeva@mail.tau.ac.il} \\\And
  Ankit Gupta\footnotemark[1] \\
  Tel Aviv University \\
  {\tt ankitgupta.iitkanpur@gmail.com} \\\AND
  Jonathan Berant \\
  Tel Aviv University, \\
  Allen Institute for AI \\
  {\tt joberant@cs.tau.ac.il} \\}

\title{Injecting Numerical Reasoning Skills into Language Models}

\date{}

\begin{document}
\maketitle

\begin{abstract}
Large pre-trained language models (LMs) are known to encode
  substantial amounts of linguistic information. However, high-level reasoning
  skills, such as numerical reasoning, are difficult to learn from a
  language-modeling objective only. Consequently, existing models for numerical reasoning
  have used specialized architectures with limited flexibility. In this work, we
  show that numerical reasoning is amenable to automatic data generation, and thus one can inject this skill into pre-trained LMs, by generating large amounts of data, and training in
  a multi-task setup. 
  We show that pre-training our model, \genbert{}, on this data, dramatically improves performance on \drop{} ($49.3 \rightarrow 72.3$ F$_1$), reaching performance that matches state-of-the-art models of comparable size, while using a simple and general-purpose encoder-decoder architecture. Moreover, \genbert{} generalizes well to math word problem datasets, while maintaining high performance on standard RC tasks. Our approach provides a general recipe for injecting skills into large pre-trained LMs, whenever the skill is amenable to automatic data augmentation.
\end{abstract}

\section{Introduction}
Recently, models trained on large amounts of data with a language modeling (LM) objective, have shown great promise in natural language processing, exhibiting surprising amounts of knowledge and information \cite{peters2018elmo, devlin2018bert, liu2019roberta, lan2019albert, petroni2019language, hewitt2019structural}. However, high-level skills, such as the ability to perform numerical reasoning over text, can be challenging to capture with a LM objective only. Consider the example in Table~\ref{table:drop_example}. To solve the first question (Q1), a model must capture the value of numbers in the text, compute their difference, and generate the tokens corresponding to the result, which generally do not appear in the input passage.

\begin{figure}\setlength{\belowcaptionskip}{-18pt}
    \centering
    \includegraphics[scale=0.39]{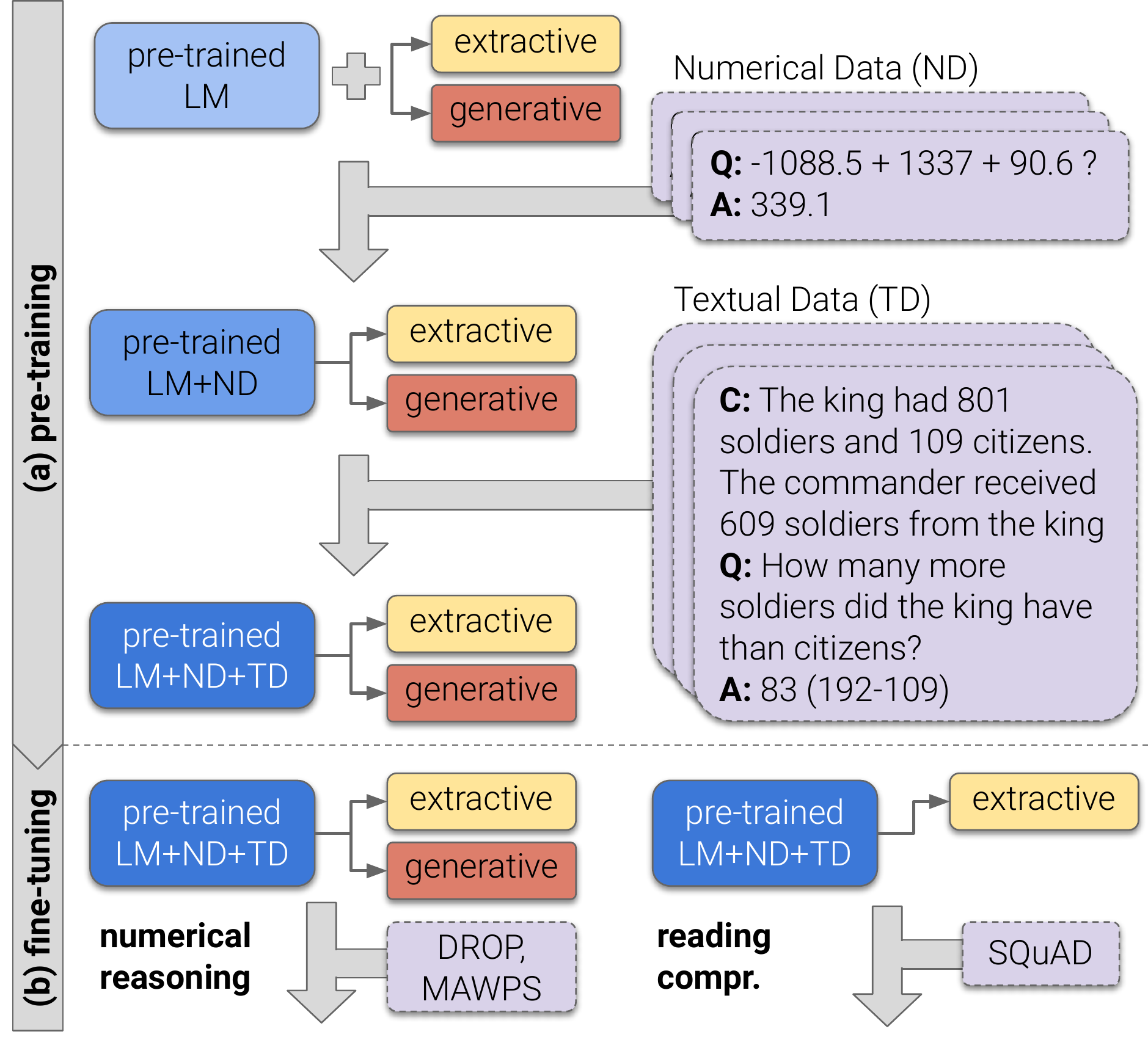}
    \caption{An overview of our approach for injecting numerical skills into a pre-trained LM. (a) We add two pre-training steps over large amounts of synthetic numerical data (ND) and textual data (TD); (b) we further fine-tune the model over either numerical reasoning datasets (\drop, \textsc{MAWPS}) or reading comprehension datasets (\squad).}
    \label{figure:intro}
\end{figure}

To make the task more manageable, state-of-the-art models have employed specialized architectures, restricting the space of possible numerical computations to a limited set. Modules were designed for counting (but only until `9') and for addition and subtraction (but of 2-3 numbers only). Such models perform well on existing datasets, such as \drop{} \cite{dua2019drop}, but do not generalize to unsupported computations, which will require modifying the model architecture.
Moreover, current models marginalize at training time over all numerical expressions that evaluate to the correct answer. Since the number of such expressions grows exponentially, scaling these approaches to arbitrary computations entails using non-differentiable operations (sampling or computing top-$K$ numerical expressions), which can lead to training difficulties.

\begin{table}
\setlength{\abovecaptionskip}{-2pt}
\setlength{\belowcaptionskip}{-15pt}
\begin{center}
\footnotesize
\begin{tabular}{p{7.0cm}}
\toprule
\textbf{Passage}: \textit{Taunton has four art galleries... \textbf{\textcolor{myblue}{Hughes/ Donahue Gallery}} founded in \textbf{\textcolor{myred}{2007}}, a local community gallery serving local Taunton artists... \textbf{\textcolor{myblue}{Art Euphoric}} founded in \textbf{\textcolor{myred}{2008}} has both visual and craft exhibits...} \\ \hline
\textbf{Q1}: How many \textbf{\textcolor{myred}{years}} \textbf{\textcolor{Green}{after}} founding of \textbf{\textcolor{myblue}{Hughes/ Donahue}} was \textbf{\textcolor{myblue}{Art Euphoric}} founded? \\
\textbf{A1}: 1 (number) \\ \hline
\textbf{Q2}: Which gallery was founded \textbf{\textcolor{Green}{later}},  \textbf{\textcolor{myblue}{Hughes/ Donahue}} or \textbf{\textcolor{myblue}{Art Euphoric}}? \\
\textbf{A2}: Art Euphoric (span) \\
\toprule
\end{tabular}
\end{center}
\caption{Example passage from \drop{}, and two questions with different answer types.}
\label{table:drop_example}
\end{table}


In this work, we propose that reasoning skills, such as numerical reasoning, are amenable to \emph{automatic data generation}. Hence, one can inject that skill directly into the model by adding additional pre-training steps, allowing the model to learn the skill in an end-to-end fashion. This results in a fully-differentiable training procedure over a standard and general-purpose architecture, where the output space can be easily controlled through the data generation procedure. 

Specifically (Figure~\ref{figure:intro}), we add to a large pre-trained LM two pre-training steps over automatically-generated synthetic data. First, we generate \emph{numerical data} of the form $3 + 4 + 11=18$. Training on this data teaches the model to compute the value of numbers from their tokens and to perform numerical operations. Second, we automatically generate question-passage pairs that require numerical reasoning using a compact grammar (\emph{textual data}). Training on this data endows the model with the ability to understand computations expressed in pseudo-natural language. 

In both pre-training steps, the model, \genbert{}, generates output numbers token-by-token. Thus, the model has a standard architecture, where an answer can either be extracted from the input question and passage or generated from a decoder. Pre-training is done in a multi-task setup with a standard LM objective, in order to avoid ``catastrophic forgetting'' \cite{kirkpatrick2017overcoming} of the linguistic information in the original LM. After pre-training, the model has sufficient language and numerical skills to be directly fine-tuned on a target numerical reasoning dataset, without resorting to specialized architectures. Augmenting more numerical skills does not require changing the model, only generating additional data.

We demonstrate the validity of our approach by a series of experiments showing that: 
\begin{enumerate}[label=(\alph*),leftmargin=*,topsep=0pt,itemsep=0pt,parsep=0pt]
    \item \genbert{} is able to solve pre-training tasks for numerical reasoning.
    \item Pre-training on these tasks provides \genbert{} with 1) skills to reach performance that matches state-of-the-art models of comparable size on \drop{} \cite{dua2019drop}, a standard numerical reasoning dataset, as well as 2) the ability to generalize to math word problem (MWP) datasets \cite{koncel2016mawps}.
    \item \genbert{} learns these numerical skills while maintaining high performance on SQuAD \cite{rajpurkar2016squad}, a standard reading comprehension dataset.
    \item Initializing models for numerical reasoning with \genbert{}'s weights improves their original performance.
\end{enumerate}



To conclude, in this work we address the problem of injecting LMs with numerical reasoning skills. Our contributions are:
\begin{itemize}[leftmargin=*,topsep=0pt,itemsep=0pt,parsep=0pt]
    \item A method for injecting skills into pre-trained LMs, given that automatic data generation is possible.
    \item \genbert{}, an architecture for pre-trained LM with generative and extractive abilities.
    \item A framework for generating numerical and textual synthetic data for numerical reasoning.

\end{itemize}


Our code and data can be downloaded from \url{https://github.com/ag1988/injecting\_numeracy}.

\section{Numerical Reasoning Over Text}
\label{sec:background}

Numerical reasoning over text (NRoT) is commonly set up as a reading
comprehension (RC) task. Given a training set of question-context-answer triples
$\{(\question_i, \context_i, a_i)\}_{i=1}^N$, the goal is to learn a function
that returns the answer $a$ to a  question $\question$ given a context
$\context$. However, in NRoT the answer generally requires to internally perform
some numerical computation using the entities and numbers in the context.
Specifically, the answer is either: (a) a span (or list of spans) from the
context $\context$ or question $\question$, or (b) a number that is the result of some computation (see examples in Table~\ref{table:drop_example}).

Two natural, yet opposing, approaches lend themselves to tackling NRoT:
(a) \emph{A symbolic approach}: a model can read the question and context, output a numerical expression and evaluate the answer with an external symbolic calculator. This approach is a particular case of semantic parsing \cite{kamath2019survey}, and was common in early NRoT datasets \cite{koncel2015parsing, roy2015solving, hosseini2014learning}. However, it suffers from several drawbacks. First, because numerical expressions are discrete and their space grows combinatorially, the model must learn to search in this space using non-differentiable operations, which are usually difficult to optimize. Second, numerical expressions are limited to numerical answers, while in \textsc{DROP} often a numerical computation is required but the final answer is a text span.
(b) \emph{A distributed approach}: have a model directly generate the answer
given $(\question,\context)$. When the answer is a text span, the model can
extract it from the input, and when the answer is a number that is not in $\question$ or
$\context$, the model must generate it. While this makes training straightforward, the model must learn to perform numerical computations from the relatively small target dataset. We empirically show in \S\ref{sec:genbert} that this leads to low performance in general.

As a compromise, most NRoT models \cite{dua2019drop, kinley2019nabert, hu2019multi, efrat2019tag} have taken a hybrid approach: they augment standard extractive QA models with specialized modules for handling a limited set of numerical computations.
We briefly describe this architecture, as it is the basis for our
model in \S\ref{sec:genbert}.

Given a question with $n_1$ tokens $\question = (q_1, \dots, q_{n_1}$) and a
context with $n_2$ tokens $\context = (c_1, \dots, c_{n_2})$, the hybrid model
first computes contextualized representations for the $n_1+n_2+3$ tokens
$\langle \texttt{[CLS] }\question\texttt{ [SEP] }\context\texttt{[SEP]} \rangle$
using a pre-trained LM, such as \textsc{BERT} \cite{devlin2018bert}:
\[
\mathbf{L} = \textbf{LM}(\question, \context).
\]
The representations $\mathbf{L}$ are then passed to multiple \emph{heads}, which
are small neural networks that estimate $p(a \mid \question, \context, h)$, that
is, the probability of the answer given the input and conditioned on a
head $h$, corresponding to a particular answer type:
\begin{itemize}[topsep=0pt,itemsep=0pt,parsep=0pt,partopsep=0pt,leftmargin=*]
    \item \emph{Context span head}: computes a distribution over all spans in the \emph{context} using a feed-forward network (FFN) $\textbf{FF}_\context(\mathbf{L})$.
    \item \emph{Question span head}: computes a distribution over spans in the \emph{question} 
    using a FFN $\textbf{FF}_\question(\mathbf{L})$.
    \item \emph{Count head}: computes a distribution over the numbers $\{0,\ldots,9\}$ using a FFN $\textbf{FF}_{\mathbf{cnt}}(\mathbf{L})$.
    \item \emph{Arithmetic head}: computes a distribution over all signed combinations of numbers in the context using a FFN $\textbf{FF}_{\mathbf{cmb}}(\mathbf{L})$ (the numbers in the context are identified in a pre-processing step).
\end{itemize}
While the first two heads are standard in extractive QA, the latter two heads are specialized and meant to handle answers that do not appear in the input.

Finally, for deciding which answer head to use for a
given input, a \emph{type} head $\textbf{FF}_{\mathbf{typ}}(\mathbf{L})$ outputs
a probability distribution $p_\text{head}(h \mid \question, \context)$ (using a FFN).
Thus the model probability for an answer is 
\[
p(a \mid \question, \context) = \sum_{h \in \text{heads}} p_\text{head}(\text{h} \mid
\context,\question) \cdot p(a \mid \context, \question, h).
\]
Training is done by enumerating all of the ways in which the answer can be obtained using all of the heads, and maximizing this marginal probability.

While existing models perform well on \drop, the aforementioned architecture is not flexible. First, the output space is severely constrained -- the model can only count up to `9',
and numerical computations are restricted to signed combinations of a few numbers.
Second, expanding the space of supported numerical computations is non-trivial, because training involves marginalizing over all expressions that lead to the correct answer. Since the space of numerical expressions grows exponentially, expanding this space quickly leads to a difficult search problem.
Third, delegating numerical computations to an external symbolic calculator leads to modeling challenges, since there could be interactions between text and numerical computation:
Consider the \drop{} question 
\nl{How many total yards did Phil Dawson throw for touchdowns?}. 
Current models handle such questions by computing a sum from numbers in the text and returning the result.
However, if the question was
\nl{Who  threw  45  total  yards  for  touchdowns?}, the model would have to compute the sum \emph{internally}, and then find the relevant span in the text.  This is impossible when the computation itself is delegated to an external calculator. Thus, training models to handle such numerical questions is desirable.


Motivated by the above arguments, we wish to push the frontier of end-to-end differentiable models for numerical
reasoning. Thus, we will automatically generate large amounts of data that endow a pre-trained LM with numerical skills.

\section{\genbert{}: A \bert{}-based Model for Generating Arbitrary Outputs}
\label{sec:genbert}

We now describe a simple \bert{}-based generative model that performs numerical computations internally, termed \genbert{}. The model combines the Transformer
encoder-decoder architecture \cite{vaswani2017attention} with a
pre-trained LM, specifically, \bert{}.

\begin{figure}
\setlength{\belowcaptionskip}{-10pt}
    \centering
    \includegraphics[scale=0.5]{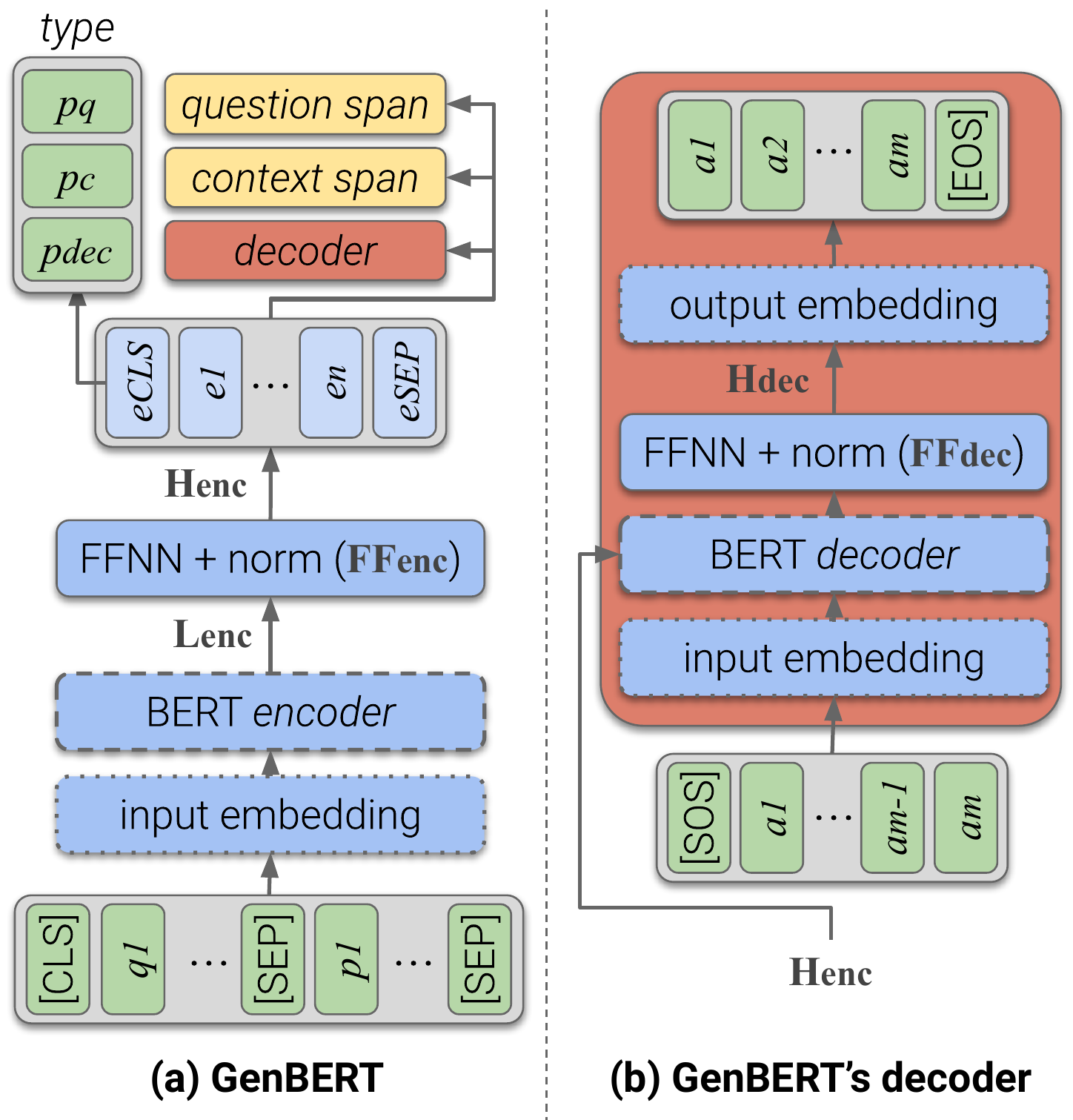}
    \caption{\genbert{}'s network architecture: (a) a high-level overview of the network, including a generative head (red), two span-extraction heads (yellow), and an answer type head. (b) a closer overview of \genbert{}'s generative head.}
    \label{figure:network_architecture}
\end{figure}

Our architecture is illustrated in Figure~\ref{figure:network_architecture}. 
Our encoder is a standard Transformer, initialized with \bert{} weights.
To also enjoy \bert{}'s representations at decoding time, 
we tie the weights of the decoder and the encoder. Because the Transformer decoder has
\emph{source attention} weights (weights for attending to the encoder
representations at decoding time) that are not present in \bert{}, we tie these
source-attention weights to the self-attention weights of the encoder (which are
tied to the self-attention weights of the decoder). This fully initializes the
Transformer model with \bert{} weights.

Since the encoder and decoder weights are tied, we make them learn
distinct representations by adding a FFN
$\mathbf{FF_{enc}}$ that transforms the encoder
contextualized representations $\mathbf{L_{enc}}$ as 
\[
\mathbf{H_{enc}} = \texttt{layer-norm}(\texttt{gelu}(W\cdot\mathbf{L_{enc}})),
\]
where $W$ is a parameter matrix \cite{hendrycks2016gelu, ba2016layer}.
Analogously, we add $\mathbf{FF_{dec}}$ to the decoder. To further distinguish
the encoder and decoder, we use distinct start and end tokens for input and
output sequences. Given $m$ answer tokens $\answer = (a_1, \dots, a_m)$, we form
an output sequence with $m+2$ tokens: $\langle \texttt{[SOS] }\answer\texttt{
  [EOS]} \rangle$. The output tokens
  are passed through the decoder and $\mathbf{FF_{dec}}$ to obtain $\mathbf{H_{dec}}$.

Finally, the probability of an answer is defined in the usual manner:
Let $\langle\answer\rangle = (a_0 \cdots a_{m+1})$ be the output
sequence. The decoder outputs the probability $p_\text{dec}(a_{i+1} \mid
a_0,..a_i,\context,\question)$, and the probability of an answer is: 
\[
p_{\text{dec}}(\langle\answer\rangle \mid \context,\question) = \prod_{i=0}^{m}
p_\text{dec}(a_{i+1} \mid a_0,..a_{i},\context,\question).
\]

 
As we have a generative model, we can remove the specialized \emph{count} and \emph{arithmetic} heads
from \S\ref{sec:background}. Thus, the type head
$\mathbf{FF_{typ}}(\mathbf{H_{enc}})$ outputs a distribution $(p_{\question},
p_{\context}, p_{\mathbf{dec}})$ over the context span, question span, and decoder heads.

To improve pre-training on the numeric data
(\S\ref{sec:pre_training_tasks}), we make two
additional modifications.

\paragraph{Digit Tokenization (DT)} 
Conventional wordpiece tokenization treats numbers no differently than any other
token. However, computing the value of numbers should be simpler
when using digits directly \cite{wallace2019numeracy}.  
Hence, we tokenize numbers digit-by-digit.
For example, a wordpiece \texttt{\#\#$d_1\cdots d_k$} where $d_i \in
\texttt{\{0,...,9\}}$ is further split into \texttt{\#\#$d_1$},
..., \texttt{\#\#$d_k$}. 
We show in \S\ref{section:pretraining_results} that this
substantially improves sample complexity when training to perform
numerical operations.

\paragraph{Random Shift (RS)}
The original Transformer uses absolute positional
embeddings for each token. 
However, in \S\ref{sec:pre_training_tasks}, we train on short inputs
such as \emph{``1086.1 - 2.54 + 343.8''}. Thus, the model can potentially over-fit and 
learn to 
perform numerical reasoning only when numbers are at the beginning of an input. To prevent this, when the
input length $n_1+n_2+3 < 512$, we shift all position IDs by a random integer in
$(0, 1, \dots, 512 - (n_1 + n_2 + 3))$.

\paragraph{Training}
For each span $(i,j)$, a span extraction head $h$ outputs its probability
$p_{h}((i,j) \mid \context,\question, h)$ of being the answer. Let $S$ be the set of spans in the input corresponding to the gold answer. 
The model loss $\mathbf{L_{\text{model}}}$ marginalizes over all ways in which the answer can be predicted:
\[
-\log\bigg(p_{\mathbf{dec}}\cdot p_{\text{dec}}(\langle\answer\rangle)  \  +  \ \sum_{\mathbf{h} \in \question,\context} p_\mathbf{h} \cdot\sum_{(i,j) \in S} p_h(i,j)\bigg),
\]
where conditionals have been dropped for brevity.

To evaluate the ability of \genbert{} to perform numerical reasoning, we
initialize it with \bert{} and fine-tune it on \drop{}. \genbert{} obtains 46.1
EM and 49.3 F$_1$, roughly 20 points
lower than prior models. Thus, we conclude that acquiring numerical reasoning
skills from \drop{} data only is difficult. 
To remedy this, we will 
automatically generate training data that will endow \genbert{} with numerical
skills before training it on \drop{}.




\section{Pre-training Tasks for Numerical Skills}
\label{sec:pre_training_tasks}
We now describe two automatically-generated datasets and the
multi-task training procedure.

\subsection{Generating Numerical Data (ND)}
\label{section:numerical_data}
Our first dataset focuses on learning numerical values expressed by tokens
and computing numerical operations, i.e., it does not involve textual content.
As such, it is easy to craft templates that correspond to various numeric operations.
We designed six such templates, described in Table~\ref{table:numerical_data}. Each template consists of an expression to evaluate and its solution. Further details on their instantiation are provided in \S\ref{section:numerical_data_details}. While the numerical operations were chosen based on \drop{}, it is trivial to extend them to other domains \cite{saxton2019analysing} with different numerical operations.


\begin{table*}[t]
\setlength{\belowcaptionskip}{-10pt}
    \scriptsize
    \centering
    \begin{tabular}{l|c|p{5cm}}
        \bf Operation & \bf Template & \bf Example instantiation \\ \hline
        signed float combination & $s_1\,f_1\,s_2\,f_2\,s_3\,f_3\,s_4\,f_4$   & 517.4 - 17484 - 10071.75 + 1013.21 \\
        min/max/avg & $o(f_1,\,f_2,\,f_3,\,f_4)$  & largest(13.42, 115.5, 72.76) \\
        $\arg\max$, $\arg\min$ & $arg(w_1\,f_1,\,w_2\,f_2,\,w_3\,f_3,\,w_4\,f_4)$  & $\arg\min$(highish 137.1, sightliness 43.2) \\
        date min/max & $dsup(d_1,\,d_2,\,d_3,\,d_4)$  & oldest(June 04, 959; 01 May 959) \\
        date difference & \text{diff in} $prd(d_1, d_2)$ & diff in days(05 April 112; June 01, 112)  \\
        percentage & $pcent\,w\,::\,w_1\,p_1\%,\,w_2\,p_2\%,\,w_3\,p_3\%,\,w_4\,p_4\%$   & percent not sunbird :: sunbird 33.2\%, defector 60.77\%, molehill 6.03\% \\
        \hline
    \end{tabular}
    \caption{Templates for generating synthetic numerical examples and the numerical operations required to answer them. \\
    \textbf{Domains} (defined in App.~\ref{section:numerical_data_details}): $s_i\in \{-,+\}$, $f_i \in \mathbb{R}^{+}$, $o \in \mathcal{O}$ : superlative words like \nl{longest}, $arg \in \text{\{$\arg\min$, $\arg\max$\}}$, $w_i \in \mathcal{W}$ : words from NTLK Words Corpus, $d_i \in \mathcal{D}$: dates until Sep 2019, $dsup \in \mathcal{DSUP}$ : superlative words like \nl{latest}, $prd \in \text{\{\nl{days}, \nl{months}, \nl{years}\}}$, $p_i \in (0,100)$, $pcent \in \text{\{\nl{percent}, \nl{percent not}\}}$.
    }  
    \label{table:numerical_data}
\end{table*}


\subsection{Generating Textual Data (TD)}
\label{section:textual_data}
Numeric data is easy to generate, since it does not contain any textual context. However, to tackle NRoT, a model needs  to comprehend how numerical operations are expressed in text that refers to events, entities and quantities. This primes us to generate \emph{textual data} from a simple grammar.


While text generation is hard in the general case, we are specifically interested in text that focuses on number manipulations. Therefore, we use the framework of \citet{hosseini2014learning}, who proposed to model math word problems with a simple structure. In their framework a \emph{world state} consists of \emph{entities}, which are objects that are being counted, and \emph{containers}, which are objects that own entities. Sentences use \emph{verb categories} to describe how the number of entities in a container changes, and thus a world state can be updated given a sentence.

Consider the textual example in Figure~\ref{figure:intro}. the \emph{entities} are soldiers and citizens, and the \emph{containers} are the king and the commander. The verbs (\nl{had} and \nl{received}) describe the entities the king holds, and how many were passed to the commander.


\begin{figure}
\setlength{\belowcaptionskip}{-15pt}
    \centering
    \includegraphics[scale=0.38]{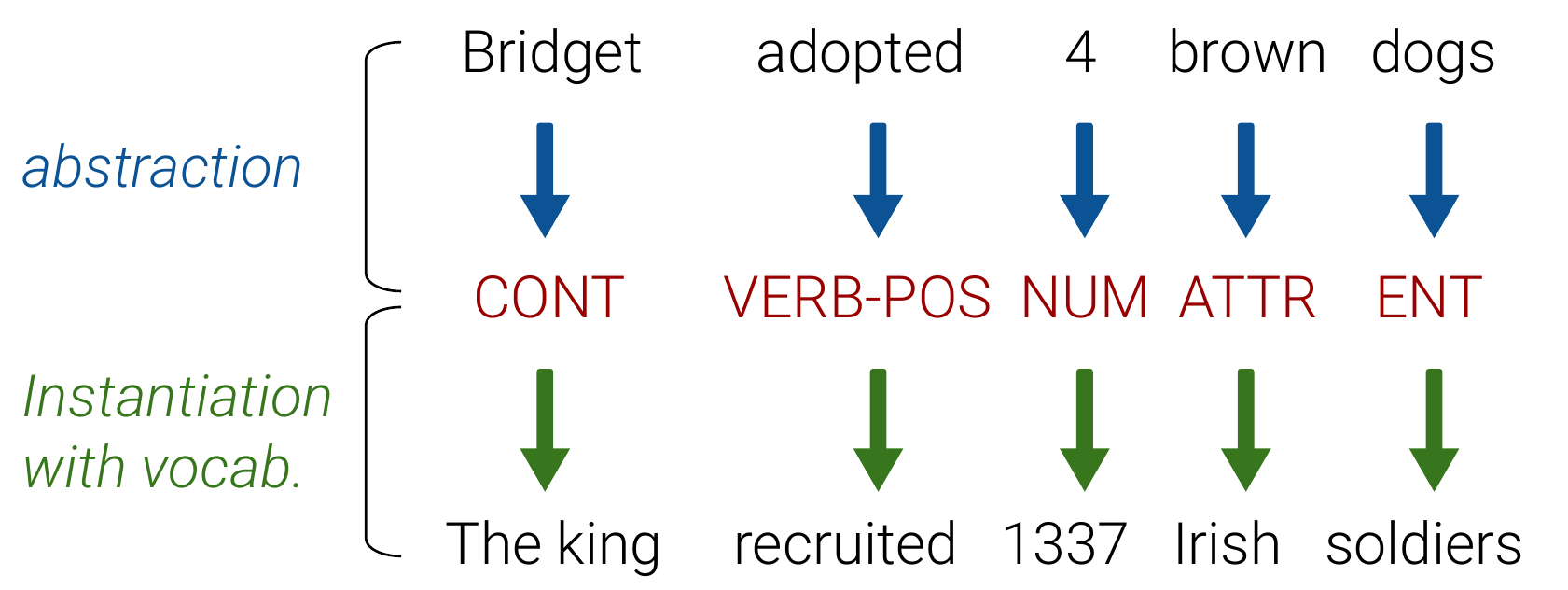}
    \caption{Template extraction and instantiation.
    A template (in red) is extracted from a MWP sentence, using categories for containers, entities, verbs, attributes and numbers, according to \citet{hosseini2014learning}. For generation, the categories are instantiated with a domain-specific vocabulary.}
    \label{figure:textual_data_generation}
\end{figure}

In this work, we use this framework to automatically generate examples. We \emph{extract templates} that describe changes in the number of entities owned by containers, and automatically \emph{generate question-context pairs} from these templates.

\paragraph{Template extraction}
To extract templates, we go over sentences from the corpus provided by \citet{hosseini2014learning}. For each sentence, we use a procedure described by~\citet{hosseini2014learning}
to abstract its tokens to the following categories:
numbers (\texttt{NUM}), entities (\texttt{ENT}), containers (\texttt{CONT}) and attributes (\texttt{ATTR}). 
In addition, verbs are abstracted to six categories, each corresponding to a different change in the number of entities owned by containers.
Thus, each template fully specifies how to update a \emph{world state}, i.e., the number of entities each container owns. 
The top part of Figure~\ref{figure:textual_data_generation} illustrates the abstraction process.
Finally, we count for each extracted template its frequency in the data, and use the top-$12$ templates for passage generation.
Details on the abstraction process, categories used, and extracted templates are in \S\ref{section:textual_data_details}.



\paragraph{Passage generation}
Using the extracted templates, we can generate sentences and maintain a world state of all containers and the number of entities they own. 
We construct a small vocabulary (<100 words) that maps categories to domain-specific words, and use the following procedure to generate passages.

We sample 3-6 templates with replacement, and instantiate them one-by-one (the bottom part of Figure~\ref{figure:textual_data_generation} illustrates instantiation).
Each template is instantiated by uniformly sampling values from the vocabulary with probability $1-p$ and from previously generated sentences with probability $p$. 
To avoid a collection of unrelated sentences, we set the probability of using previously used values to $p=0.7$.
An example passage is shown in Table~\ref{table:synthetic_textual_example}. 

\paragraph{Question generation}
After generating a passage, the world state holds information about all containers in the passage and the number of entities they hold. In Table~\ref{table:synthetic_textual_example}, the state will include the number of families and rebels of different nationalities in each container (the commander, the householder, and the countries).
Based on this world state, numerical reasoning questions can be asked.

To create questions, we craft 13 question templates that are instantiated with objects from the world state. The questions teach the model to track events and perform numeric and discrete operations. Examples for generated questions are shown in Table~\ref{table:synthetic_textual_example}, where answers are computed from the world state.
Overall, we create 13 question templates for 7 different ``skills", provided in \S\ref{section:textual_data_details}.

\begin{table}[t]\setlength{\abovecaptionskip}{-2pt}\setlength{\belowcaptionskip}{-15pt}
\begin{center}
\footnotesize
\begin{tabular}{p{7.2cm}}
\toprule
\textbf{P}: The commander recruited 1949 Polish families in Spain. The householder recruited 1996 Japanese families in Spain. There were 10913 white rebels and 77 Chinese families in Spain. 6641 British soldiers, 476 asian rebels, and 338 Germans families were recruited in Russia. \\ \hline
\textbf{Q}: How many Japanese families were in Spain? \\
\textbf{A}: 1996 \\ 
\textbf{Q}: How many more Japanese families were in Spain than Polish families? \\
\textbf{A}: 47 (1996-1949) \\ 
\textbf{Q}: How many families of Spain were not Polish families? \\
\textbf{A}: 2073 (4022-1949) \\
\toprule
\end{tabular}
\end{center}
\caption{An example synthetic passage (P) and questions. Questions (Q) were generated from templates and answers (A) were calculated based on the world state.}
\label{table:synthetic_textual_example}
\end{table}

\subsection{Training \genbert{} on Synthetic Data}

For pre-training on ND, we generated 1M examples for training and 10K for validation. For TD, we generated 2.5M examples for training and 10K for validation. For both synthetic datasets, we used the \genbert{} model loss, $\mathbf{L_{\text{model}}}$, from \S\ref{sec:genbert}. To ensure that the model does not lose its language understanding abilities, we employ a multi-task setup, and include a standard \emph{masked LM} objective from \bert{}. Specifically, given a masked token sequence $\langle\mathbf{m}\rangle$,
we compute the contextualized representations, $\mathbf{L_{enc}}$ and
pass them through a feed-forward network $\mathbf{FF_{\text{mlm}}}$. For each masked index $i$, it outputs the probability $p(a_i \mid i,\langle\mathbf{m}\rangle)$ of the original token $a_i$. The MLM loss is computed as 
$$\mathbf{L_{\text{mlm}}}(\langle\mathbf{m}\rangle) = \mathrm{mean}_{i \in \text{masked}} -\log(p(a_i \mid i, \langle\mathbf{m}\rangle)).$$
Details about the MLM data are in \S\ref{sec:mlm_data}.

During training, we sample mini-batches from the respective datasets, and minimize the weighted sum of the losses. Concretely, while pre-training on ND and TD, we sample mini-batches $X_{\text{ND}}$, $X_{\text{TD}}$ and $X_{\text{MLM}}$ and optimize the objective 
\[
\mathbf{L_{\text{model}}}(X_{\text{ND}})\ +\ \mathbf{L_{\text{model}}}(X_{\text{TD}})\ + \ \lambda\cdot \mathbf{L_{\text{mlm}}}(X_{\text{MLM}}).
\]

\section{Experimental Evaluation}

We now evaluate our two pre-training steps and their applicability for numerical reasoning tasks. 
We consider the following variants, aiming to investigate the contributions of ND and TD, the importance of MLM loss, and techniques like DT and RS. In all cases, we initialize \genbert{} with \bert{}-base-uncased, use DT and RS, and include the MLM loss, \textit{except where noted}:
\begin{itemize}[topsep=0pt, itemsep=1pt, leftmargin=.1in, parsep=2pt]
    \item \textsc{GenBERT\ssc{+ND}}: trained on numerical data.
    \item \textsc{GenBERT\ssc{+ND-LM}}: trained on ND without the additional MLM loss.
    \item \textsc{GenBERT\ssc{+ND-LM-DT}}: trained on ND using wordpiece tokenization, without the MLM loss.
    \item \textsc{GenBERT\ssc{+ND-LM-RS}}: trained on ND without MLM loss and random shift (RS).
    \item \textsc{GenBERT\ssc{+TD}}: trained on textual data (TD). 
    \item \textsc{GenBERT\ssc{+ND+TD}}: \textsc{GenBERT\ssc{+ND}} trained on both ND and TD.
\end{itemize}

\subsection{Pre-training Performance}
 \label{section:pretraining_results}
 
\begin{figure}
\setlength{\belowcaptionskip}{-15pt}
    \centering
    \includegraphics[scale=0.39]{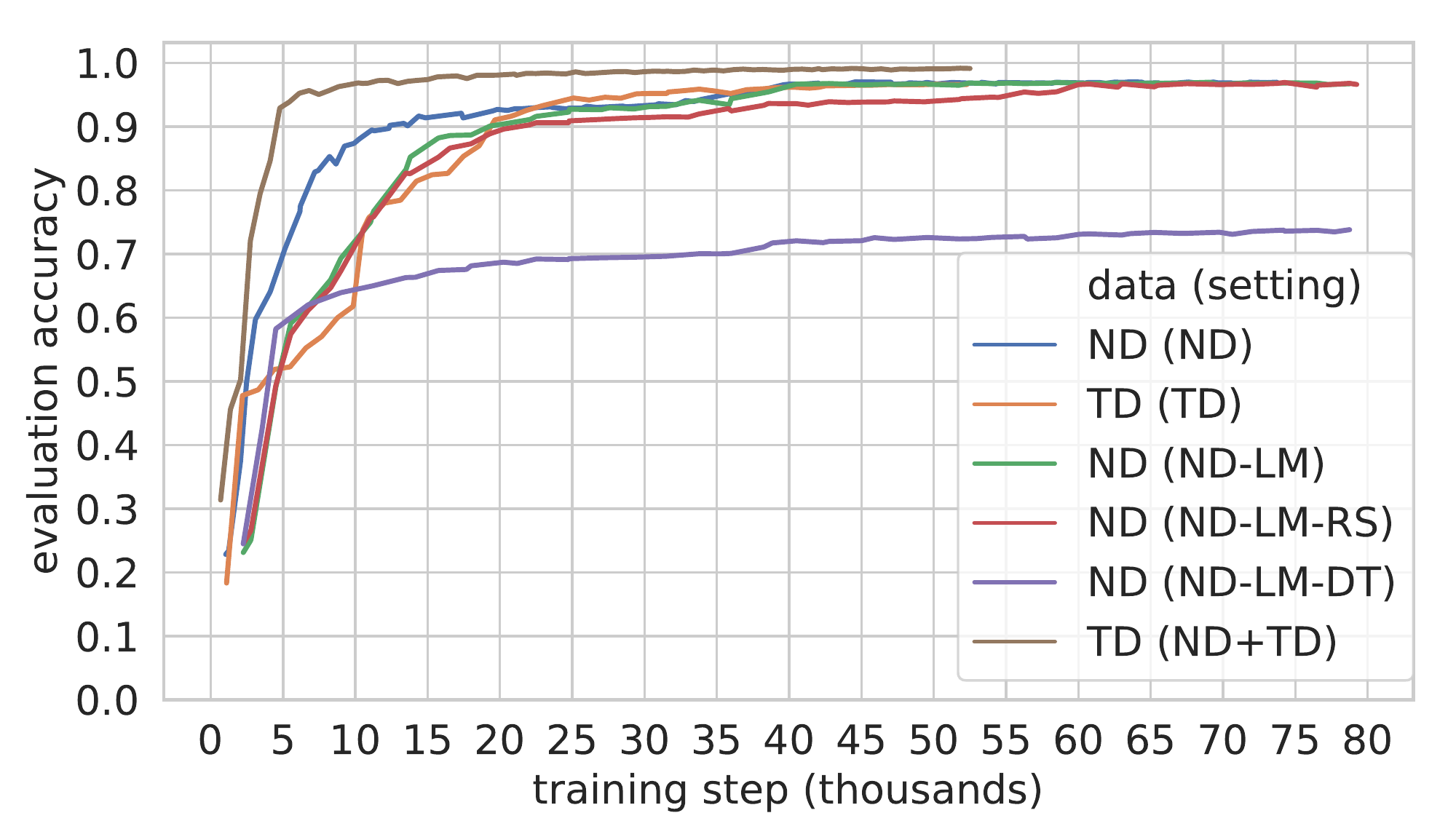}
    \caption{Progression of eval accuracy (EM) of \genbert{}, for different pre-training settings listed in \S\ref{section:pretraining_results}.}
    \label{figure:performance_pretraining}
\end{figure}


We first ask whether the pre-training procedure allows \genbert{} to absorb the intended numerical skills. We observe that across various settings (ND, TD, ND+TD), \genbert{} consistently achieves more than $96\%$ accuracy in predicting the correct solution for both ND and TD. Thus, we conclude that indeed a pre-trained LM can learn the designed skills from generated data.

Figure~\ref{figure:performance_pretraining} shows the learning curves of \genbert{} for the different variants. 
Note that in \textsc{ND-LM-DT} the model does \emph{not} learn to solve the numerical data task. This demonstrates the utility of using DT over conventional wordpieces. The lower sample complexity in the case of \textsc{ND+TD} compared to the only-\textsc{TD} can be attributed to the fact that ND and TD share some numeric skills and hence a model already trained on ND converges faster on TD compared to \genbert{}.

\subsection{Numerical Reasoning Performance}
\label{section:numerical_reasoning_performance}

After successfully injecting \genbert{} with numeric skills, we test \genbert{} guided by the following questions:
\begin{enumerate}[label=(\alph*),leftmargin=*,topsep=0pt,itemsep=0pt,parsep=0pt]
    \item Are the injected skills robust and generalize to NRoT datasets like \drop?
    \item Are the new skills learned at the expense of the model's ability to understand language?
    \item Can the pre-trained weights be used with architectures other than \genbert{}?
\end{enumerate}
For (a), we fine-tune \genbert{} on \drop{} and further evaluate on MWP in a zero-shot setup .
For (b), we evaluate \genbert{} on a RC task that does not involve numerical reasoning, namely, \squad{} \cite{rajpurkar2016squad}. For (c), we use \genbert{} encoder as a drop-in replacement for \bert{} on two other architectures.


\paragraph{Results on \drop{}}
We report results of \genbert{} initialized by \bert{}-base and leave pre-training a larger model 
for future work. We compare \genbert{} to 
\mtmsn{} \cite{hu2019multi} initialized with \bert{}-base, as \mtmsn{} initialized with \bert{}-large is a state-of-the-art model on \drop{}.\footnote{Per ACL policy, we compare to models that were made public 3 months prior to submission.}

Table~\ref{table:drop_performance} presents fine-tuning results on \drop{}. 
Without pre-training, \genbert{} performs poorly compared to current state of the art models like \mtmsn, reporting an EM of only 46.1. Pre-training on each of the numerical data (ND) and textual data (TD) improves performance dramatically to 64.7 EM and 64.4 EM, respectively. Moreover, pre-training on both ND and TD leads to a performance of 68.8 EM, on par with MTMSN's 68.2 EM. This demonstrates that the skills that \genbert{} learns from ND and TD are complementary.
In addition, the lower performance of \textsc{GenBERT\ssc{+ND-LM}} and \textsc{GenBERT\ssc{+ND-LM-RS}} shows the importance of including the MLM loss and the utility of RS for short inputs.

\begin{table}[t]\setlength{\belowcaptionskip}{-5pt}
    \footnotesize
    \centering
    \begin{tabular}{l|c|c|c|c|}
         & \multicolumn{2}{c|}{Development} & \multicolumn{2}{c|}{Test} \\ 
         & EM & F$_1$ & EM & F$_1$ \\ \hline
         \textsc{GenBERT} & 46.1 & 49.3 & - & -\\
         \textsc{GenBERT\ssc{+ND-LM-RS}} & 61.5 & 65.4 & - & -\\
         \textsc{GenBERT\ssc{+ND-LM}} & 63.8 & 67.2 & -& -\\
         \textsc{GenBERT\ssc{+ND}} & 64.7 & 68.2 & -& -\\
         \textsc{GenBERT\ssc{+TD}} & 64.4 & 67.8 & -& -\\
         \textsc{GenBERT\ssc{+ND+TD}} & \textbf{68.8} & 72.3 & \textbf{68.6} & \textbf{72.4} \\
         \hline\hline
         \nabert & 63.0 & 66.0 & 61.6 & 65.1 \\
         \mtmsn \textsubscript{\textsc{BASE}} & 68.2 & \textbf{72.8} & -& -\\
    \end{tabular}
    \caption{Performance of \genbert{} and comparable models on the development and test sets of \drop{}.}
    \label{table:drop_performance}
\end{table}

Breaking down performance by answer type (Table~\ref{table:performance_per_answer_type}) highlights several points. First, pre-training on ND and TD improves performance mostly due to number answer types, as expected.
Second, \textsc{GenBERT\ssc{+ND+TD}} outperforms \mtmsn \textsubscript{\textsc{BASE}} on questions whose answer is \emph{a span}. We argue a probable cause for this are span questions that require performing a numerical computation internally, as explained in \S\ref{sec:background}.
Third, \mtmsn \textsubscript{\textsc{BASE}} substantially outperforms \genbert{} on questions whose answer is a list of non-contiguous spans. This is expected, as \mtmsn{} has a specialized head and procedure for handling such questions, while build on a simpler and more standard RC architecture.

\begin{table}[t]\setlength{\belowcaptionskip}{-10pt}
    \footnotesize
    \centering
    \begin{tabular}{l|c|c|c|c|}
         & number & span & date & spans \\ \hline
         \textsc{GenBERT} & 42.3 & 67.3 & 47.5 & 21.1 \\
         \textsc{GenBERT\ssc{+ND}} & 70.5 & 71.0 & 54.5 & 24.2 \\
         \textsc{GenBERT\ssc{+TD}} & 69.2 & 72.6 & 55.2 & 22.0 \\
         \textsc{GenBERT\ssc{+ND+TD}} & \bf 75.2 & \bf 74.5 & \bf 56.4 & 24.2 \\
         \hline\hline
         \nabert & 67.8 & 69.2 & 39.8 & 22.4 \\
         \mtmsn \textsubscript{\textsc{BASE}} & \bf 75.0 & 71.3 & 44.2 & \bf 53.4 \\
    \end{tabular}
    \caption{F$_1$ scores on DROP development per answer type.}
    \label{table:performance_per_answer_type}
\end{table}

\comment{
\begin{table}[t]
    \footnotesize
    \centering
    \begin{tabular}{l|c|c|c|}
         & super. & comp. & pct. \\
          & (1,237) & (2,627) & (1,556) \\\hline
         \textsc{GenBERT} & 51.0 & 38.9 & 69.2 \\
         \textsc{GenBERT\ssc{+ND}} & 61.0 & 65.8 & 79.4  \\
         \textsc{GenBERT\ssc{+TD}} & 62.8 & 65.1 & 77.7  \\
         \textsc{GenBERT\ssc{+ND+TD}} & \bf 64.3 & \bf 72.2 & 82.6 \\
         \hline\hline
         \nabert & 60.9 & 63.5 & 72.9  \\
         \mtmsn \textsubscript{\textsc{BASE}} & 62.2 & \bf 72.2 & \bf 84.1  \\
    \end{tabular}
    \caption{Performance on DROP development for questions containing superlatives (super.), comparatives (comp.), and percentages (pct.). The number of examples in each subset is shown at the top.}
    \label{table:performance_per_question_type}
\end{table}
}

\paragraph{Generalization to MWP (zero-shot)}  
The MAWPS repository is a collection of math word problem (MWP) datasets \cite{koncel2016mawps}. To test the models on skills they were trained on, we
picked datasets with addition and subtraction problems, and filtered out examples with other operations (e.g., multiplication and division). All models that were fine-tuned on \drop{} were evaluated in a zero-shot setup on 395 examples from \addsub{} \cite{hosseini2014learning}, 321 from \singleop{} \cite{roy2015reasoning}, and 305 from \singleeq{} \cite{koncel2015parsing}.

Results are shown in Table~\ref{table:mwp_performance}. 
Overall, \textsc{GenBERT\ssc{+ND+TD}} dramatically improves performance compared to \genbert{}. \textsc{GenBERT\ssc{+ND}} performs much better than \textsc{GenBERT\ssc{+TD}}, demonstrating the utility of ND when the context is short. Last, \mtmsn{} outperforms \textsc{GenBERT\ssc{+ND+TD}}. However, \mtmsn{} uses a specialized architecture for addition and subtraction, suitable when calculations are done outside of the model. \genbert{}, on the other hand, is a general-purpose generative model, that can also return span answers when the computation is done internally.

\begin{figure}
\setlength{\abovecaptionskip}{-2pt}
    \centering
    \includegraphics[scale=0.39]{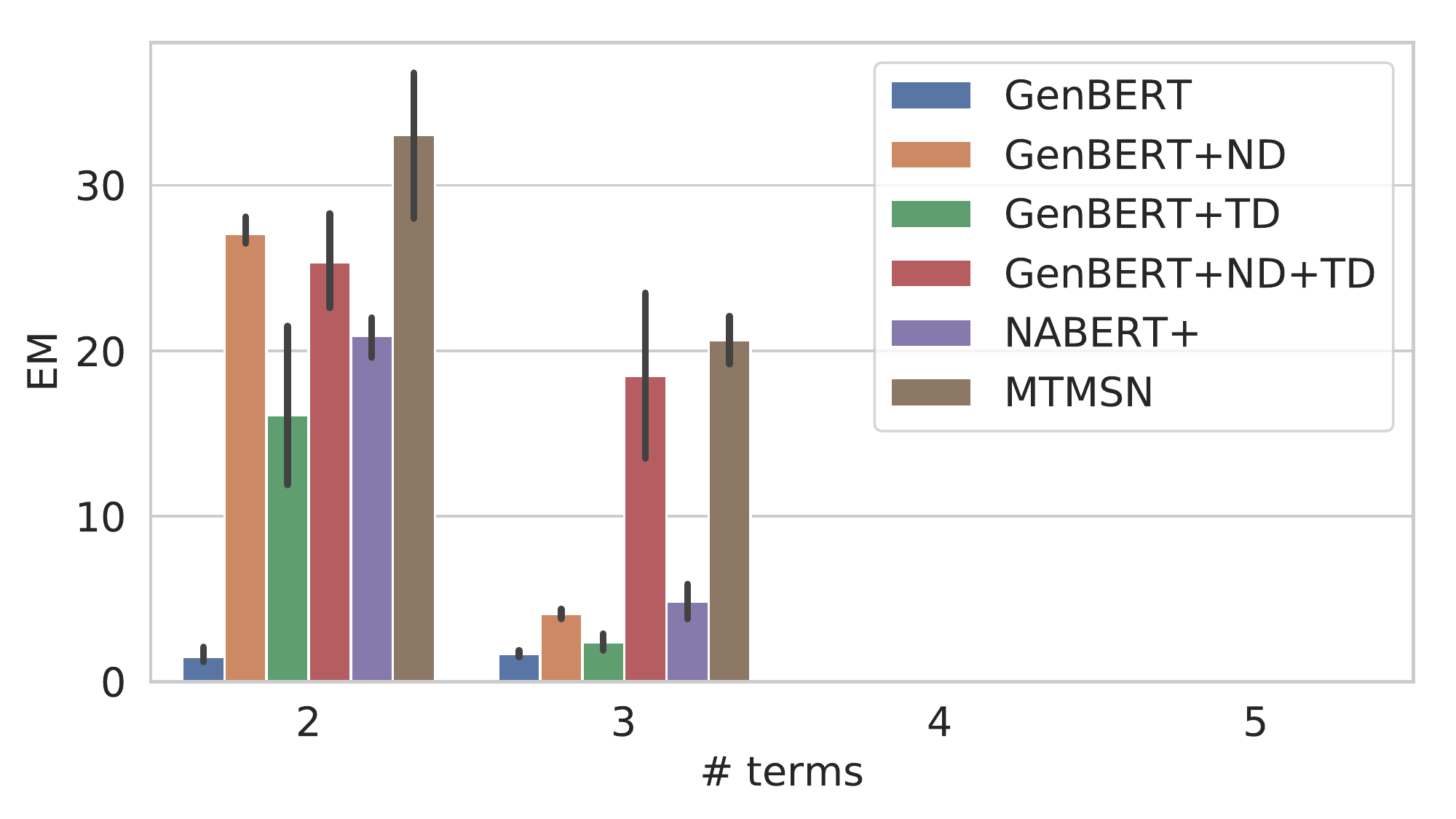}
    \caption{Breakdown of model accuracy (EM) by the number of terms in the arithmetic expression, for the MWP datasets \addsub{}, \singleop{} and \singleeq{}.}
    \label{figure:mwp_performance_by_num_terms}
\end{figure}

Next, we break down performance by the number of terms in the arithmetic expression (Figure~\ref{figure:mwp_performance_by_num_terms}). The plot shows that all models struggle to generalize to more complex problems, and completely fail when the calculation involves more than 3 terms.
Interestingly, the drop in performance of \textsc{GenBERT\ssc{+ND+TD}} between 2 and 3 terms is significantly smaller than that of \textsc{GenBERT\ssc{+ND}} and \textsc{GenBERT\ssc{+TD}}. This suggests that both ND and TD are useful for improving robustness.



\begin{table}[t]\setlength{\belowcaptionskip}{-10pt}
    \footnotesize
    \centering
    \begin{tabular}{l|c|c|c}
          & \addsub & \singleop & \singleeq \\ \hline
         \textsc{GenBERT} & 2 & 1.2 & 1.3 \\
         \textsc{GenBERT\ssc{+ND}} & 22.8 & 26.5 & 23 \\
         \textsc{GenBERT\ssc{+TD}} & 10.4 & 21.5 & 12.1 \\
         \textsc{GenBERT\ssc{+ND+TD}} & 22.8 & \bf 28.3 & 22.3 \\
         \hline\hline
         \nabert & 19.2 & 19.6 & 17.4 \\
         \mtmsn \textsubscript{\textsc{BASE}} & \bf 32.2 & 28 & \bf 32.5 \\
    \end{tabular}
    \caption{EM on MWP datasets.}
    \label{table:mwp_performance}
\end{table}

\paragraph{Error analysis}  
To understand the limitations of our method, we analyze the errors of \textsc{GenBERT\ssc{+ND+TD}} on the development set of \drop{}, excluding questions with a multi-span answer which are not supported by the model.
We sample 100 random examples for which \textsc{GenBERT\ssc{+ND+TD}} fails to predict the correct answer and manually analyze the types of questions and mistakes done by the model.

We find that in almost half of the cases (43\%), the example requires reasoning skills that are either not covered by the pre-training tasks (e.g. sorting), or not numerical. 
Another common case (23\%) is inaccurate predictions, such as spans that are too long and numbers with partial digit match to the gold answer.
We note that many of these errors can be addressed by extending the pre-training tasks to cover additional numerical skills and a larger number range. We leave such extensions for future work.
Further details and example failure cases are provided in \S\ref{sec:error_analysis}.

\subsection{Reading Comprehension Performance}
Having shown that our models successfully learned to perform NRoT, we investigate if this improvement comes at the expense of performance on RC datasets. We initialize the RC model from 
\citet{devlin2018bert} with \genbert{} weights (encoder only) and fine-tune it on \squad{} v1. As shown in
Table~\ref{table:squad_performance}, the performance of  \textsc{GenBERT\ssc{+ND+TD}} is almost identical to the original \bert{}. Moreover, \textsc{GenBERT\ssc{+ND-LM}} reported a loss of 3 EM points  highlighting the importance of using the MLM loss.


\begin{table}[t]\setlength{\belowcaptionskip}{-10pt}
    \footnotesize
    \centering
    \begin{tabular}{l|c|c|}
         & EM & F$_1$ \\ \hline
         \textsc{BERT} & \bf 81.1 & \bf 88.6 \\
         \textsc{GenBERT\ssc{+ND-LM}} & 78.1 & 85.8 \\
         \textsc{GenBERT\ssc{+ND}} & 80.7 & 88.1 \\
         \textsc{GenBERT\ssc{+TD}} & 80.7 & 88.2 \\
         \textsc{GenBERT\ssc{+ND+TD}} & \bf 81.3 & \bf 88.6
    \end{tabular}
    \caption{Performance on SQuAD v1 development set. Scores for \textsc{BERT} are using wordpiece tokenization.
    }
    \label{table:squad_performance}
\end{table}

\subsection{\genbert{} With Other Architectures} 
To further establish the utility of \genbert{}, we used the weights of \textsc{GenBERT\ssc{+ND+TD}} to initialize the encoder of \nabert{} and \textsc{MS-TAG}, a recent multi-span tagging model of \citet{efrat2019tag}. 
Fine-tuning on \drop{} 
shows an improvement of $\sim$2 EM points compared to the originally reported performance: $63.0 \rightarrow 65.1$ EM for \nabert{}, and $67.3 \rightarrow 69.3$ EM for \textsc{MS-TAG}. This shows that \genbert{} can be used as a drop-in replacement for \bert{}, when numerical reasoning is needed.


To summarize, we have empirically shown that one 
can inject numerical reasoning skills into a pre-trained LM, resulting in good performance on \drop{}, generalization to MWP, while maintaining high performance on standard RC datasets. Moreover, the resulting weights can be used for initializing numerical reasoning models.
\section{Related Work}





Most NRoT models designed for \drop{} are extractive QA models augmented with specialized modules (\S\ref{sec:background}).
Two recent work \cite{andor2019giving, chen2020neural} take a more symbolic approach and output a symbolic program augmented with operations over text.
In our work, numerical computations are latent and performed internally by the model.

A related line of work has been analyzing the mathematical reasoning abilities of neural models over text \cite{wallace2019numeracy, rozen2019diversify, ravichander2019equate}, and on arithmetic problems \cite{saxton2019analysing, amini2019mathqa,lample2020deep}.

Designing pre-training tasks to teach LMs additional skills has been applied by \citet{huang2019unicoder}, who designed cross-lingual pre-training tasks to teach better mappings between languages, and \citet{lee2019latent}, who introduced the Inverse Cloze Task to pre-train an information retriever. 

\section{Conclusions}
Large pre-trained LMs lack high-level skills such as numerical reasoning. Consequently, current models that perform numerical reasoning over a pre-trained LM resorted to customized modules with limited flexibility.
In this work, we propose a general method for injecting additional skills into LMs, assuming automatic data generation is possible.
We apply our approach to the task of numerical reasoning over text, using a general-purpose model called \genbert{}, and a simple framework for generating large amounts of synthetic examples.
Our experiments demonstrate the effectiveness of our method, showing that \genbert{} successfully learns the numerical skills, and performs on par with state-of-the-art NRoT models of the same size.

\section*{Acknowledgments}
We thank Daniel Andor and Thang Luong for helpful discussions, and Shimi Salant for constructive suggestions.
This research was partially supported by
The Israel Science Foundation grant 942/16, 
The Yandex Initiative for Machine Learning, and the European Research Council (ERC) under the European Union Horizons 2020 research and innovation programme (grant ERC DELPHI 802800).

\bibliography{all}
\bibliographystyle{acl_natbib}

\clearpage

\appendix

\section{Supplemental Material}
\label{sec:supplemental}

\subsection{Synthetic Numerical Data Generation}
\label{section:numerical_data_details}

We briefly describe the numerical templates, providing the details missing from Table~\ref{table:numerical_data}. In all cases, integers are sampled from $\{0,\ldots,20K\}$, and split into disjoint train and development sets to assure generalization.

\begin{itemize}[leftmargin=*,topsep=0pt,itemsep=-0.2em,topsep=0pt,parsep=0pt]
\item \emph{signed float combination} : Random signed combinations of up to 4 floats. Floats are sampled from the set of floats with two decimal places.

\item \emph{min/max/avg} :
We sample 2-4 floats and apply a \texttt{min}, \texttt{max}, \texttt{avg}, operation by sampling a word from the set $\mathcal{O}$ = \{\nl{longest}, \nl{last}, \nl{highest}, \nl{largest}, \nl{most}, \nl{shortest}, \nl{first}, \nl{smallest}, \nl{lowest}, \nl{least}, \nl{average}\}.

\item \emph{$\arg\max$, $\arg\min$}: 
We sample word-float pairs, where words are sampled from $\mathcal{W}$: words in the NLTK Words Corpus\footnote{\texttt{https://www.nltk.org/}} having at most 2 wordpieces, and floats are sampled as above.

\item \emph{date max/min} : Same as \emph{min/max/avg} above, but for dates. Dates are sampled from $\mathcal{D}$: the set of dates until Sep 2019. The operator word is sampled from $\mathcal{DSUP}$ = \{\nl{last}, \nl{latest}, \nl{most recent}, \nl{youngest}, \nl{first}, \nl{earliest}, \nl{oldest}, \nl{least recent}\} and mapped to $\min$ or $\max$.

\item \emph{date difference} : This teaches our model to perform date arithmetic in days, months and years.

\item \emph{percentage} : We teach our model to perform $100-x$ operations in the context of percentages. Given a number of arguments, we sample a percentage split using a flat Dirichlet distribution.
\end{itemize}

\subsection{Synthetic Textual Data Generation}
\label{section:textual_data_details}

\subsubsection{Sentence template extraction}
To extract sentence templates, we abstract the text of math word problems from the corpus published by \citet{hosseini2014learning}.
Going over examples, we split the problem text into sentences\footnote{Using the Spacy library \texttt{http://spacy.io/}}, and abstract the tokens of each sentence independently. Tokens are abstracted according to the framework into numbers (\texttt{NUM}), verb categories (\texttt{VERB}), entities (\texttt{ENT}), containers (\texttt{CONT}) and attributes (\texttt{ATTR}).

To have a better control over the generation process, we extend the framework of \citet{hosseini2014learning} to support two container types - agent (\texttt{AGT}) and environment (\texttt{ENV}). Agents are objects which actively collect and drop entities, for example a person or an organization. Environments are passive containers, such as places or time periods. In addition, we introduce two-level containers to express inclusion relation between containers. For instance, if 3 submarines anchor near the city of Devonport, then they also anchor near the country of England.

The 12 most common extracted sentence templates, which were used for generating synthetic data, are provided in Table~\ref{table:synthetic_passages}.

\subsubsection{Template instantiation}
Sentence templates are instantiated with a small vocabulary, that map categories into words.
In this work, we construct two domain-specific small-world vocabularies, about history and the National Football League. 
The vocabularies are available in a json format in \texttt{\url{https://github.com/ag1988/injecting_numeracy}}.

\subsubsection{Question templates}
The 13 question templates for 7 different skills are provided in Table~\ref{table:synthetic_textual_questions}.

\begin{table*}[t]
    \footnotesize
    \centering
    \begin{tabular}{p{15.0cm}}
        \bf Template\\ 
        \toprule
        \texttt{CONT-1-AGT} \texttt{VERB-1-*} \texttt{NUM-1} \texttt{ATTR-1} \texttt{ENT-1} . \\
        \texttt{CONT-1-AGT} \texttt{VERB-1-POS} \texttt{NUM-1} \texttt{ATTR-1} \texttt{ENT-1} and \texttt{CONT-2-AGT} \texttt{VERB-1-POS} \texttt{NUM-2} \texttt{ATTR-1} \texttt{ENT-1} . \\
        \texttt{CONT-1-AGT} \texttt{VERB-1-POS} \texttt{NUM-1} \texttt{ATTR-1} \texttt{ENT-1} and \texttt{NUM-2} \texttt{ATTR-2} \texttt{ENT-2} . \\
        \texttt{CONT-1-AGT} \texttt{VERB-1-POS} \texttt{NUM-1} \texttt{ATTR-1} \texttt{ENT-1} , but \texttt{VERB-2-NEG} \texttt{NUM-2} \texttt{ATTR-2} \texttt{ENT-2} . \\
        \texttt{CONT-1-AGT} \texttt{VERB-1-POS} \texttt{NUM-1} \texttt{ATTR-1} \texttt{ENT-1} in \texttt{ATTR-2} \texttt{CONT-2-ENV} . \\
        \texttt{CONT-1-AGT} \texttt{VERB-1-NEG} \texttt{NUM-1} of the \texttt{ATTR-1} \texttt{ENT-1} . \\
        \texttt{CONT-1-AGT} had \texttt{NUM-1} \texttt{ATTR-1} \texttt{ENT-1} , \texttt{CONT-2-AGT} had \texttt{NUM-2} \texttt{ATTR-1} \texttt{ENT-1} , and \texttt{CONT-3-AGT} had \texttt{NUM-3} \texttt{ATTR-1} \texttt{ENT-1} . \\
        \texttt{NUM-1} \texttt{ATTR-1} \texttt{ENT-1} , \texttt{NUM-2} \texttt{ATTR-2} \texttt{ENT-2} , and \texttt{NUM-3} \texttt{ATTR-3} \texttt{ENT-3} were \texttt{VERB-1-POS} in \texttt{ATTR-4} \texttt{CONT-1-ENV} . \\
        There were \texttt{NUM-1} \texttt{ATTR-1} \texttt{ENT-1} and \texttt{NUM-2} \texttt{ATTR-2} \texttt{ENT-2} in \texttt{ATTR-3} \texttt{CONT-1-ENV} . \\
        There were \texttt{NUM-1} \texttt{ATTR-1} \texttt{ENT-1} in \texttt{ATTR-2} \texttt{CONT-1-ENV} . \\
        \texttt{CONT-1-AGT} \texttt{VERB-1-NEGTRN} \texttt{NUM-1} \texttt{ATTR-1} \texttt{ENT-1} to \texttt{CONT-2-AGT} . \\
        \texttt{CONT-1-AGT} \texttt{VERB-1-POSTRN} \texttt{NUM-1} \texttt{ATTR-1} \texttt{ENT-1} from \texttt{CONT-2-AGT} . \\
    \end{tabular}
    \caption{Sentence templates for synthetic textual examples.}
    \label{table:synthetic_passages}
\end{table*}

\begin{table*}[t]
    \footnotesize
    \centering
    \begin{tabular}{lp{11.2cm}}
        \bf Reasoning & \bf Templates \\ 
        \toprule
        Selection & How many \texttt{ATTR-1} \texttt{ENT-1} were in \texttt{CONT-1-ENV}? \\
         & How many \texttt{ATTR-1} \texttt{ENT-1} did \texttt{CONT-1-AGT} \texttt{VERB-POS}? \\ \hline
        Intra-entity difference & How many more \texttt{ATTR-1} \texttt{ENT-1} were in \texttt{CONT-1-ENV} than \texttt{ATTR-2} \texttt{ENT-2} ? \\
         & How many more \texttt{ATTR-1} \texttt{ENT-1} did \texttt{CONT-1-AGT} have than \texttt{ATTR-2} \texttt{ENT-2} ? \\ \hline
        Intra-entity subset & How many \texttt{ENT-1} of \texttt{CONT-1} were \texttt{ATTR-1} \texttt{ENT-1} ? \\
         & How many \texttt{ENT-1} of \texttt{CONT-1} were not \texttt{ATTR-1} \texttt{ENT-1} ? \\ \hline
        Inter-entity comparison & Were there \{more | less\} \texttt{ATTR-1} \texttt{ENT-1} in \texttt{CONT-1-ENV} or in \texttt{CONT-2-ENV} ? \\
         & Who had \{more | less\} \texttt{ATTR-1} \texttt{ENT-1}, \texttt{CONT-1-AGT} or \texttt{CONT-2-AGT} ? \\ \hline
        Inter-entity superlative & Who had the \{highest | lowest\} number of \texttt{ATTR-1} \texttt{ENT-1} in total ?\\ \hline
        Intra-entity superlative & What was the \{highest | lowest\} number of \texttt{ATTR-1} \texttt{ENT-1} \texttt{VERB-POS} in \texttt{CONT-1-ENV} ?\\
         & What is the \{highest | lowest\} number of \texttt{ATTR-1} \texttt{ENT-1} \texttt{CONT-1-AGT} \texttt{VERB-POS} ?\\ \hline
         Inter-entity sum & How many \texttt{ATTR-1} \texttt{ENT-1} were in \texttt{CONT-1-ENV} (, \texttt{CONT-*-ENV}) and \texttt{CONT-2-ENV} \{in total | combined\} ?\\
          & How many \texttt{ATTR-1} \texttt{ENT-1} did \texttt{CONT-1-ENV} (, \texttt{CONT-*-ENV}) and \texttt{CONT-2-ENV} have \{in total | combined\} ?\\ \hline
    \end{tabular}
    \caption{Templates for questions about generated synthetic passages, testing for numerical reasoning. The template placeholders are filled-in with values from the world state obtained after generating the synthetic passage.}
    \label{table:synthetic_textual_questions}
\end{table*}

\subsection{Data for Masked LM task}\label{sec:mlm_data}
For creating the training data for the masked LM task (\S~\ref{section:pretraining_results}) we took the pages from English Wikipedia whose lowercased title containing a string in \{\textit{season, economy, demographics, conquest, war, battle, uprising, rebellion, insurgency, conflict, crisis, revolution, military history, mutiny, regiment, revolt, geography, raids, insurrection, invasion, feud, siege, campaign, expedition, succession, coup, university}\}. This resulted in 156K full pages. In the remaining pages, paras with < 15 numbers were discarded. Pages were tokenized using DT (\S~\ref{sec:genbert}) and chunked into 512-token sequences. Following \citet{devlin2018bert}, each token was masked with probability 0.15 with no more than $65$ masks per sample. This gave us 0.7M samples.


\subsection{Experimental Setup}\label{sec:exp_setup}
For all our experiments, we used an older version of Hugging Face's Transformers library \cite{Wolf2019HuggingFacesTS} and provide our training hyperparameters in Table~\ref{table:hyperparams}.
\begin{table}[t]\setlength{\belowcaptionskip}{-5pt}
    \scriptsize
    \centering
    \begin{tabular}{l|c|c|c|c|c|}
         & \multicolumn{3}{c|}{pre-training} & \multicolumn{2}{c|}{finetuning} \\
         \hline
         & lr & bsz & epochs & lr & bsz \\ \hline
         \textsc{GenBERT} & - & - & - & 3e-5 & 16 \\
         \textsc{GenBERT\ssc{+ND}} & 6e-5 & 800 & 60 & 3e-5 & 16 \\
         \textsc{GenBERT\ssc{+ND-LM}} & 4e-5 & 440 & 60 & 3e-5 & 16 \\
         \textsc{GenBERT\ssc{+ND-LM-DT}} & 4e-5 & 440 & 60 & - & - \\
         \textsc{GenBERT\ssc{+ND-LM-RS}} & 4e-5 & 440 & 60 & 3e-5 & 16 \\
         \textsc{GenBERT\ssc{+TD}} & 1e-5 & 240 & 5 & 3e-5 & 14 \\
         \textsc{GenBERT\ssc{+ND+TD}} & 1e-5 & 240 & 5 & 3e-5 & 14 \\
    \end{tabular}
    \caption{Hyperparameters used for pre-training \genbert{} and finetuning it on \drop{}. lr=leaning rate, bsz=train batch size. Common params: seed=42, optimizer=Bert-Adam, linear-lr-warm-up=0.1, num epochs for finetuning=30, weight-decay=0.01, max-grad-norm=1.0.}
    \label{table:hyperparams}
\end{table}

\subsection{\textsc{GenBERT\ssc{+ND+TD}} Error Analysis}\label{sec:error_analysis}

Table~\ref{table:error_analysis} summarizes the main failure types of \textsc{GenBERT}\ssc{+ND+TD} on 100 random examples from the development set of \drop{}, excluding questions with a multi-span answer.

\begin{table*}[t]
    \footnotesize
    \centering
    \begin{tabular}{p{3cm}|p{12cm}}
         Error category & Example \\
         \toprule
         Counting & \textbf{q}: \textit{How many people were the heads of the business?} \\
         Sorting & \textbf{q}: \textit{Which nationality was the fourth largest?} \\
         Complex calculation & \textbf{q}: \textit{How many percent of people were either Black Hispanic, of Sub-Saharan African origin, or of West Indian or Afro-Caribbean American origin?} \\
         Complex semantics & \textbf{q}: \textit{By how many points did the Pistons lose their closest game?} \\
         Not numerical & \textbf{q}: \textit{Who defeated the Kievan Rus at the Battle of the Alta River?} \\
         \hline \hline
         \multirow{3}{3cm}{Longer span} & \textbf{q}: \textit{Where there more people in the peninsula pre-war or at the time of the first census?} \\
         & \textbf{a}: pre-war \\
         & \textbf{p}: pre-war population \\
         \hline
         \multirow{3}{3cm}{Shorter span} & \textbf{q}: \textit{Was the life expectancy in 2015 higher for males or females?} \\
         & \textbf{a}: females \\
         & \textbf{p}: female \\
         \hline
         \multirow{3}{3cm}{Imprecise number prediction} & \textbf{q}: \textit{How many more estimated Chinese Americans lived in California compared to Massachusetts?} \\
         & \textbf{a}: 1130100 \\
         & \textbf{p}: 110100 \\
         \hline
    \end{tabular}
    \caption{Error categories of \textsc{GenBERT\ssc{+ND+TD}} on the development set of \drop{}, based on a manual error analysis of 85 random examples. The upper part shows categories which are not not covered by our pre-training tasks or do not require numerical skills. The lower part shows categories of inaccurate model predictions. The letters \textbf{q}, \textbf{a} and \textbf{p} denote the question, gold answer and model prediction, respectively.}
    \label{table:error_analysis}
\end{table*}

\end{document}